\documentclass[3p,times,procedia]{elsarticle}
\usepackage{ecrc}
\usepackage[bookmarks=false]{hyperref}
    \hypersetup{colorlinks,
      linkcolor=blue,
      citecolor=blue,
      urlcolor=blue}
\usepackage{amsmath,amssymb,amsfonts}

\volume{00}

\firstpage{1}

\journalname{Procedia Computer Science}

\runauth{Author name}

\jid{procs}

\usepackage{amssymb}

\usepackage[figuresright]{rotating}

\begin{document}
\begin{frontmatter}



\dochead{Proceedings of the 2024 International Conference on Biomimetic Intelligence and Robotics}

\title{Head Pose Estimation and 3D Neural Surface Reconstruction via Monocular Camera in situ for Navigation and Safe Insertion into Natural Openings}


\author[a]{Ruijie Tang\fnref{label2}}
\author[a]{Beilei Cui\fnref{label2}}
\author[a,b,c,d]{Hongliang Ren\corref{cor1}}

\address[a]{Department of Electronic Engineering, The Chinese University of Hong Kong, Hong Kong, China}
\address[b]{Department of Biomedical Engineering, National University of Singapore, Singapore}
\address[c]{Shun Hing Institute of Advanced Engineering, The Chinese University of Hong Kong, Hong Kong, China}
\address[d]{Shenzhen Research Institute, The Chinese University of Hong Kong, Shenzhen, China}

\begin{abstract}
As the significance of simulation in medical care and intervention continues to grow, it is anticipated that a simplified and low-cost platform can be set up to execute personalized diagnoses and treatments. 3D Slicer can not only perform medical image analysis and visualization but can also provide surgical navigation and surgical planning functions. In this paper, we have chosen 3D Slicer is our base platform, and monocular cameras are used as sensors. Then, We used the neural radiance fields (NeRF) algorithm to complete the 3D model reconstruction of the human head. We compared the accuracy of the NeRF algorithm in generating 3D human head scenes and utilized the MarchingCube algorithm to generate corresponding 3D mesh models. The individual’s head pose, obtained through single-camera vision, is transmitted in real-time to the scene created within 3D Slicer. The demonstrations presented in this paper include real-time synchronization of transformations between the human head model in the 3D Slicer scene and the detected head posture. Additionally, we tested a scene where a tool, marked with an ArUco Maker tracked by a single camera, synchronously points to the real-time transformation of the head posture. These demos indicate that our methodology can provide a feasible real-time simulation platform for nasopharyngeal swab collection or intubation.
\end{abstract}

\cortext[cor1]{Corresponding author.}
\fntext[label2]{R. Tang and B. Cui are co-first authors.}
\end{frontmatter}

\email{hlren@ee.cuhk.edu.hk}

\vspace*{-6pt}

\section{Introduction}
\label{main}

Digital Twin has enormous potential to improve disease treatment, optimize medical procedures, and provide better care~\cite{khan2023digital,li2023rethinking,tang2024data,yao2023rnn}.  One typical application is surgical simulation and planning~\cite{bjelland2022toward,shu2023twin}. By creating a digital twin model of the patient, doctors can conduct virtual simulation and planning before the surgery. Digital Twins, combined with sensor technology, can monitor and collect patient physiological data in real-time and predict disease states and health trends through data analysis and modeling. It can also be used in medical education, providing practical experience and training opportunities for students and doctors. However, setting up a Digital Twin system is expensive, requiring an optical tracker and VR equipment ~\cite{lee2023robotic,du2023digital}.

The COVID-19 epidemic has caused significant obstacles in various areas of human life. During the COVID-19 pandemic, digital twins provided a method for working in isolation~\cite{gao2021progress,courtney2021using}. Among them, we discuss the workings of two critical medical interventions. One is nasopharyngeal swab sampling; the other is intubation. Manual sampling puts medical personnel in danger of infection. Robotic sampling can potentially reduce this danger to a bare minimum, but typical robots have concerns with safety, cost, and control complexity that make them unsuitable for large-scale deployment~\cite{hu2021design}. For intubation, it is challenging for a robot to manipulate the tube into the nasal cavity~\cite{cheng2018intubot}. The greatest challenge lies in inserting at an appropriate angle that can make subsequent actions smoother. Otherwise, it would take multiple adjustments of the angle to succeed. Under the digital twin system, completing these two medical tasks requires generating a personalized 3D head model and monitoring the posture of the head in real-time.

In order to avoid using devices such as calipers or 3D scanners, we need a simple, low-cost method to accomplish high-quality 3D reconstruction~\cite{cui2024endodac,cui2024surgical}. Photogrammetry, a method highly esteemed among image-based 3D reconstruction techniques, can generate an accurate and dense 3D point cloud from a real-life setting using multiple images captured from diverse perspectives~\cite{remondino2006image}. In recent years, the method of 3D scene reconstruction by NeRF using a monocular camera dataset has emerged as a novel approach~\cite{mildenhall2021nerf,wang2021nerf}. NeRF assigns a fully connected network, otherwise known as a neural radiance field, to duplicate the initial scene views, utilizing a rendering loss. Compared to the photogrammetry 3D reconstruction approach, NeRF can generate less noisy results when the object is texture-less, highly reflective, and transparent~\cite{remondino2023critical,huang2024endo}. 

Despite head pose estimation being a relatively mature algorithm, it remains lacking as an independent Python module in 3D Slicer and in the development of downstream medical applications. The head posture estimation methods based on monocular cameras have received widespread attention due to their simplicity and extensive usability~\cite{ji20023d,murphy2008head}. Feature-based methods estimate the pose by detecting key facial or head points, such as eyes, mouth, and nose. They use the position and movement of these points to infer the angles of the head's pose. Primary techniques include those based on traditional machine learning and optimization algorithms (e.g., Artificial Potential Field models, Particle Filters)~\cite{fanelli2011real,vatahska2007feature}. The development of head pose estimation methods heavily relies on extensive annotated datasets and the creation of evaluation metrics. Some commonly used datasets include 300W, CelebA, and AFLW, each offering abundant facial pose annotations~\cite{andrea2023deep}. Evaluation metrics typically encompass angular error measurements (such as Euclidean distance) and pose classification accuracy, intended to gauge the precision and stability of the estimates.

3D Slicer, with its essential medical image analysis function, has numerous expandable modules suited for specific medical applications. For instance, an image segmentation tool can be integrated into the Slicer~\cite{liu2023samm}. To deal with morphological variability with 3D data, SlicerMorph offers similar segmentation functions but develops shape variation for better visualisation~\cite{rolfe2021slicermorph}. A significant portion of the medical applications of 3D Slicer are based on medical image segmentation. Downstream applications can include, for example, hip surgical planning~\cite{mandolini2022comparison}. As for 3D reconstruction of the head and the aforementioned medical applications (such as intubation and nasal swabbing), the current 3D Slicer does not yet possess the relevant modules.

Therefore, we propose a simplified Digital Twin platform based on 3D Slicer in this article. This method can avoid the need for expensive optical tracking devices and can complete model reconstruction, real-time identification, and feature-matching functions using algorithms. In response to the issues brought about by COVID-19, our primary focus is on medical applications pertinent to the human head. Our platform can be utilized whether it is nasopharyngeal swab sampling, unmanned nasal intubation, or an unmanned endoscopic examination of the nasal and oral cavities.

\section{Method}

\begin{figure}[ht]
    \centering
    \includegraphics[width=0.6\linewidth, trim=0 0 0 0]{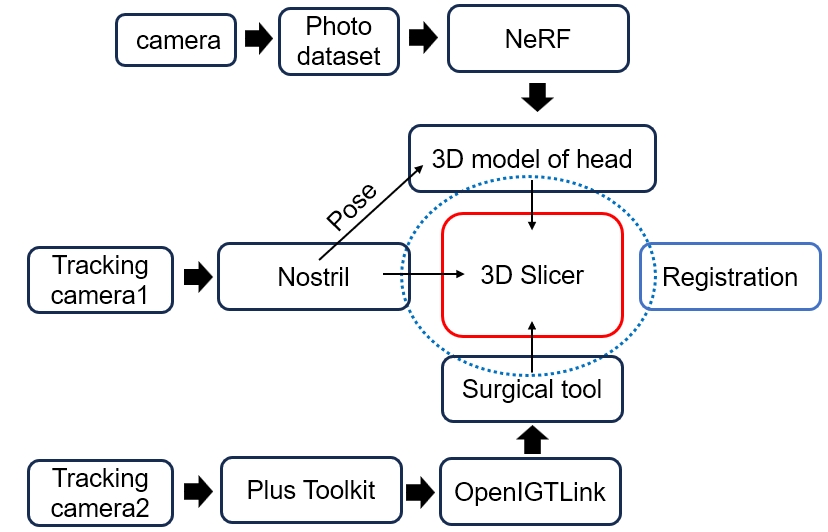}
    \caption{The proposed framework of head digital twin simulator based on 3D Slicer.}
    \label{fig:workflow}
\end{figure}
Figure~\ref{fig:workflow} depicts the entire workflow of the head digital twin simulator, and we can divide the entire method into three sub-modules. The first module generates a 3D mesh model of the target patient's head, utilizing multiple-angle photographs taken by a camera as the dataset. After that, the NeRF algorithm is applied to generate the model. The second module involves a camera to aim at the target person's head and capture the real-time head posture (roll, yaw, pitch). The task of the third module is registration, which involves aligning real-time facial images, 3D head models, and potential surgical tools. In this section, we will proceed to detail the components within the modules.

\subsection{3D model reconstruction}

We chose the NeRF to represent a 3D scene~\cite{mildenhall2021nerf}. The scene is expressed as an implicit function. A continuous volumetric function can map a 3D position $x$ and a corresponding viewing direction $d$ to color $c$ and volume density $\sigma$. The formulation is described as follows:

\begin{equation}
(\mathbf{c}, \sigma)=F_{\boldsymbol{\Theta}}\left(\gamma_{L_x}(\mathbf{x}), \gamma_{L_d}(\mathbf{d})\right),
\end{equation}

$F_{\boldsymbol{\Theta}}$ is an abbreviation for Multilayer Perceptron (MLP) with chosen parameters $\boldsymbol{\Theta}$ while $\gamma_{L}(\cdot )$ stands for positional encoding, which translates input vectors into a spectrum with a higher frequency. From the center of the camera, a line or 'ray' is projected towards the viewing direction $d_{p}$. This procedure is expressed as $\mathbf{r}_{\mathbf{p}}(t)=\mathbf{o}+t \mathbf{d}_{\mathbf{p}}$, where any pixel at image p defines the coordinate of the image. After this, we evaluate a range of sorted distances $\left\{ t^{(i)}\right\}_{i=1}^{D}$ between the predetermined near and far plane. The color, denoted as $c^{(i)}$, and density indexed as $\sigma^{(i)}$ are then computed at every individual sampling point $\mathbf{r}_{\mathbf{p}}(t^{(i)})$ using the MLP $F_{\boldsymbol{\Theta}}$. This process is then repeated to estimate the color of each pixel.

\begin{equation}
\hat{\mathbf{c}}_{\mathbf{p}}=\hat{\mathbf{c}}\left(\mathbf{r}_{\mathbf{p}}\right)=\sum_{i=1}^D T^{(i)}\left(1-\exp \left(-\sigma^{(i)} \delta^{(i)}\right)\right) \mathbf{c}^{(i)},
\end{equation}

where $\delta^{(i)} = t^{(i+1)} - t^{(i)}$ and $T^{(i)}=\exp \left(-\sum_{j=1}^{i-1} \sigma^{(j)} \delta^{(j)}\right)$. We proposed to utilize Deblur-NeRF~\cite{ma2022deblur}, a NeRF implementation that explicitly models the blurring process to solve the blur issues within images as our baseline in this paper.

\begin{figure}[ht]
    \centering
    \includegraphics[width=0.8\linewidth, trim=0 0 0 0]{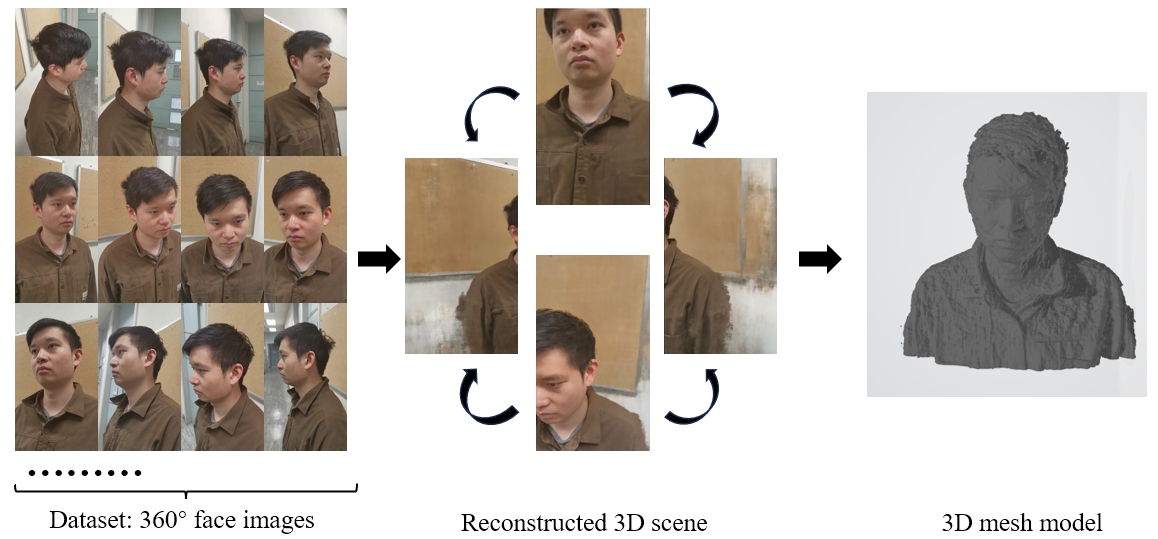}
    \caption{The workflow and results of 3D model reconstruction.}
    \label{fig:r8}
\end{figure}

Finally, with the trained MLP on our provided scene, we utilize the Marching Cubes (MC) algorithm to convert the implicit representation into an explicit representation with voxels to obtain the 3D mesh model of human head~\cite{lorensen1998marching}. The 3D mesh model is presented in Figure~\ref{fig:r8}.

\subsection{Head pose estimation}
Firstly, we need to locate the position of the head within the frame along with the associated facial landmarks. To locate the face, a Caffe model of OpenCV's DNN module is used~\cite{jia2014caffe}. Since capturing frontal facial images in real scenarios is rather challenging, a step usually added after face detection is known as ``face alignment." This procedure involves detecting key points within facial images and marking out specific areas such as eyebrows, corners of eyes, corners of mouth, etc. This critical point detection behavior is called ``Facial landmark localization." Then, a facial landmark detector is applied to find 68 landmarks based on five pre-trained datasets. The output is an array with coordinates of facial landmarks. After that, the pose estimation algorithm is applied. The algorithm starts with defining camera coordinates ($X, Y, Z$) and world coordinates ($U, V, W$). The transformation equation between two coordinates can be expressed as: 
\begin{equation}
\label{eqn1}
\begin{bmatrix}X \\Y \\Z\end{bmatrix}=\begin{bmatrix}\bold{R}  & |  & \bold{t}\end{bmatrix}\begin{bmatrix}U \\V \\W \\1\end{bmatrix}
\end{equation}
,where $\bold{R}$ is the rotation matrix and $\bold{t}$ is the translation matrix. Then, the 2D points on the image plane($x,y$) can be calculated as:  
\begin{equation}
\label{eqn2}
\begin{bmatrix}x \\y \\1\end{bmatrix}=s\begin{bmatrix}f_{x}   &0  & c_{x}\\ 0 & f_{y} & c_{y}\\ 0 & 0 &1\end{bmatrix}\begin{bmatrix}X \\Y \\Z \end{bmatrix}
\end{equation}
, where $s$ is the unknown scale factor, $f_{x}$ and $f_{y}$ are focal lengths for related directions, and ($c_{x}, c_{y}$) represents the optical center. To find out $\bold{R}$ and $\bold{t}$, we need to combine the Eqn.~\ref{eqn1} and Eqn.~\ref{eqn2}. The ($x,y$) can be obtained by feature points based on the facial landmark detector. The transformation matrix is relatively accurate when the error of the projected points of 3D points ($U, V, W$) in the world coordinate between ($x,y$) is minimized. The Levenberg-Marquardt optimization algorithm can solve $\bold{R}$ and $\bold{t}$. Using 3D Slicer and the internal Python interactor, images are captured using the monocular camera and calculated with an online solver (solvePnP). After that, the image is passed to the 3D Slicer as a volume. The transformation matrix is converted to angles (yaw, roll, pitch) and positions (x,y).

\subsection{Registration} 

In Section 2.3, we mentioned that the real-time face image also serves as a volume in the scene creation within the 3D Slicer. In that case, we maintain the eyes of the three-dimensional head model and the face image on the same plane. The scale of the image could also be fixed with the distance between the left eye left corner and the right eye right corner. The coordinates of the nose tip, left eye left corner, and right eye right corner are passed to the mesh head model in real time. Thus, We no longer need to register the relative positions of the model and the face image on the same plane. We can avoid the registration of The rotation angle for the same reason.

The second part involves registering potential surgical tools and the head model in the scene. Our system used a stick marked by an ArUco marker to simulate a surgical tool needle. Then, the position of the marker is recorded through another tracking camera. The video stream in the tracking camera is sent to 3D Slicer via the Plus toolkit and serves as a modality. We selected the nose and the two eyes as registration points to match the needle created in the scene. The final registration error is at 3.56 mm. 

\section{Results and Demonstration}

\begin{figure}[ht]
    \centering
    \includegraphics[width=0.8\linewidth, trim=0 0 0 0]{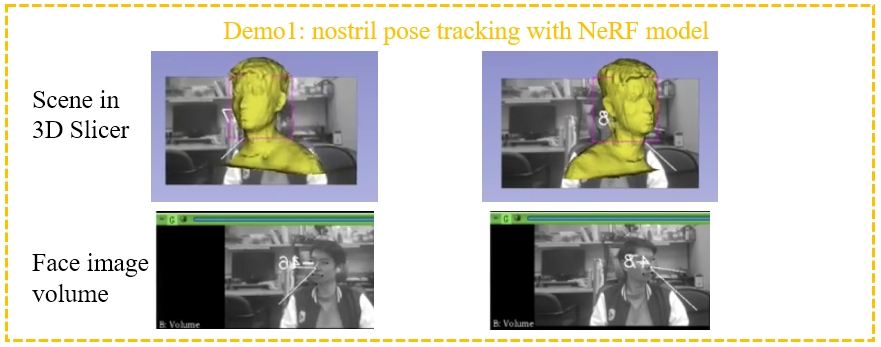}
    \caption{The demonstration of head pose tracking and simulation in 3D Slicer.}
    \label{fig:r6}
\end{figure}

\begin{table}[!h]
\caption{Quantitative results on our proposed human head dataset for different NeRF baseline methods.The results are the higher the better for PSNR and SSIM and the lower the better for LPIPS.}
\fontsize{8}{10}\selectfont
\centering
\resizebox{0.5\textwidth}{!}{
\begin{tabular}{c|ccc}
\hline 
Method & PSNR $\uparrow$ & SSIM $\uparrow$ & LPIPS $\downarrow$ \\ \hline 
NeRF~\cite{mildenhall2021nerf} & 29.91 & .8310 & .1862 \\ 
Deblur-NeRF~\cite{ma2022deblur} & 31.26 &.8436 & .2069 \\
\hline 
\end{tabular}} 
\label{tab:main}
\end{table}

\begin{figure}[ht]
    \centering
    \includegraphics[width=0.8\linewidth, trim=0 0 0 0]{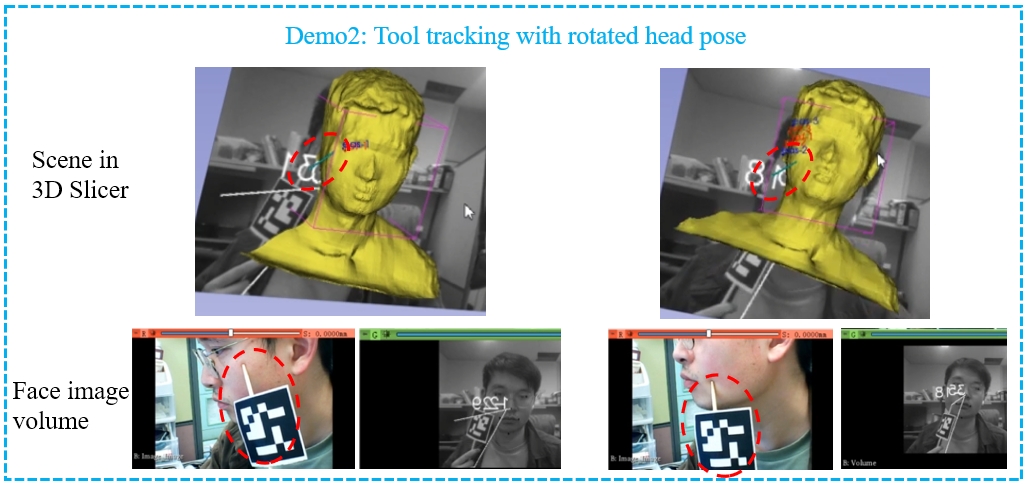}
    \caption{The demonstration of head pose tracking and simulation in 3D Slicer.}
    \label{fig:r7}
\end{figure}
In the demonstration section, we will show two possibilities. One is the simple head rotation and tracking. The other is a scenario with simulated surgical tools and real-time head tracking to demonstrate the scene created in the Slicer. In Figure~\ref{fig:r6}, we demonstrate the match between the real-time face image and the 3D model regarding yaw angle. It can be observed that the real-time face image and the model rotation angle can maintain good consistency, whether it is a clockwise or counter-clockwise rotation.
The two can similarly maintain good precision regarding real-time consistency at the roll angle. The only exception is the pitch direction. When the head is lowered or raised, the facial feature points captured by the camera become unstable, resulting in instability in the models within the scene. When the head moves out of the camera's capture screen, the 3D head remains stationary. We also evaluated the performance for different NeRF baseline methods on our human head dataset, and the results are shown in Table~\ref{tab:main}. We can notice that Deblur-NeRF obtains higher PSNR and SSIM while original NeRF results in lower LPIPS.

In Figure~\ref{fig:r7}, we tested whether the real-time rotation of the head and the direction of the tool can remain consistent when a simulated surgical tool is introduced. First, we demonstrated that when the face remains relatively still, the actual and simulated scenes can be in the same position (left-eye corner). Then, when the posture of the head deviates, a clockwise roll angle deviation occurs. We found that the tool points to the face clamp's lower half in the scene. The tool is also at the cheek in the simulated scene, slightly higher than the actual situation. Therefore, we can see that this system's primary source of error comes from the instability generated when tracking the face to calculate the angles. Additionally, when the system processes multiple video stream inputs, real-time performance may also be impaired due to hardware limitations.

\section{Conclusion}

3D Slicer can reconstruct 3D models through MRI, CT, and other images. However, traditional medical images cannot reconstruct the target's appearance. Moreover, building a digital twin system is very expensive. Therefore, in this paper, we take the head detection as the target and, in combination with 3D Slicer and NeRF, propose a pipeline to achieve 3D reconstruction and registration of the appearance. This complete pipeline's proposal will aid surgeries performed through the nasal cavity.

\section*{Acknowledgements}
This work was supported by Hong Kong Research Grants Council (RGC) Collaborative Research Fund (CRF) C4026-21GF, General Research Fund (GRF) 14203323, GRF 14216022, GRF 14211420, NSFC/RGC Joint Research Scheme N\_CUHK420/22, Shenzhen-Hong Kong-Macau Technology Research Programme (Type C) STIC Grant 202108233000303, and the key project 2021B1515120035 (B.02.21.00101) of the Regional Joint Fund Project of the Basic and Applied Research Fund of Guangdong Province. 




\bibliographystyle{cas-model2-names}

\bibliography{ref}

\clearpage

\normalMode

\end{document}